\def\year{2020}\relax
\documentclass[letterpaper]{article} 
\usepackage{aaai20}  
\usepackage{times}  
\usepackage{helvet} 
\usepackage{courier}  
\usepackage[hyphens]{url}  
\usepackage{graphicx} 
\usepackage{multirow}
\usepackage{amsmath}
\usepackage{ulem}
\def\UrlFont{\rm}  
\usepackage{graphicx}  
\title{Privacy Enhanced Multimodal Neural Representations for Emotion Recognition }
\author{Mimansa Jaiswal, Emily Mower Provost\\University of Michigan\\
\{mimansa, emilykmp\}@umich.edu} 

\makeatletter
\renewcommand\section{\@startsection {section}{1}{\z@}%
                                    {-0.65ex \@plus -0.2ex \@minus -.2ex}%
                                    {0.2ex \@plus.12ex}%
                                    {\normalfont\Large\bfseries}}
 \renewcommand\subsection{\@startsection{subsection}{2}{\z@}%
                                        {-0.5ex\@plus -0.1ex \@minus -.2ex}%
                                        {0.5ex \@plus 0.12ex}%
                                        {\normalfont\large\bfseries}}
 \renewcommand\subsubsection{\@startsection{subsubsection}{2}{\z@}%
                                        {-0.5ex\@plus -0.1ex \@minus -.2ex}%
                                        {0.5ex \@plus 0.12ex}%
                                        {\normalfont\large\bfseries}}
\makeatother

\begin{document}

\maketitle

\begin{abstract}
Many mobile applications and virtual conversational agents now aim to recognize and adapt to emotions. To enable this, data are transmitted from users' devices and stored on central servers.  Yet, these data contain sensitive information that could be used by mobile applications without user's consent or, maliciously, by an eavesdropping adversary. In this work, we show how multimodal representations trained for a primary task, here emotion recognition, can unintentionally leak demographic information, which could override a selected opt-out option by the user. We analyze how this leakage differs in representations obtained from textual, acoustic, and multimodal data. We use an adversarial learning paradigm to unlearn the private information present in a representation and investigate the effect of varying the strength of the adversarial component on the primary task and on the privacy metric, defined here as the inability of an attacker to predict specific demographic information. We evaluate this paradigm on multiple datasets and show that we can improve the privacy metric
while not significantly impacting the performance on the primary task. To the best of our knowledge, this is the first work to analyze how the privacy metric differs across modalities and how multiple privacy concerns can be tackled while still maintaining performance on emotion recognition.
\end{abstract}


\section{Introduction}
Virtual conversational agents strive to emulate human-like interaction to have more naturally flowing conversation~\cite{metcalf2019mirroring}.
These agents often employ models that classify aspects of communication, including the classification of the emotional content of speech.
The resulting predictions can then be used to bias 
response generation. Emotion classification is also used in mobile and web applications to identify heightened risk of suicidal ideation or mood fluctuations~\cite{Khorram2018}, for the purpose of tracking or intervention. Data are sent from users' devices, including mobile applications~\cite{Khorram2018} and Alexa or Google home devices~\cite{piersol2019pre}, and are stored on central servers for analysis.

However, data transmitted from users' devices are vulnerable to data hacking and re-identification~\cite{barbaro2006face}. Eavesdroppers can use these data for identification of an individual and to gain access to sensitive information.
A way to counter this issue in data collected by mobile or smart home applications is to generate a data representation on the device and then to transfer that representation to the server for additional processing. 
The benefit is that these representations can increase privacy by partially obfuscating the actual content of the conversation~\cite{bengio2013representation}. However, they still contain sensitive demographic information.

The implications of sensitive information leakage is profound: research has shown that discrimination occurs across variables of age, race, and gender in hiring, policing and credit ratings~\cite{hajian2012study}.~\cite{abadi2016deep} showed how adding random noise to aggregated dataset or individual samples can ensure defense against privacy attacks.
But, previous research has shown that privacy induced using additional noise can often be exploited if the adversary has access to the network used to generate anonymity~\cite{kifer2011no}.
Therefore to ensure robustness, we consider a scenario of the attacker having access to the same embedding sub-network to generate the representations that will be used to train its attack network.

In this work, we focus on privacy in the context of emotion recognition.  Emotion recognition provides an important test case for emotion production varies significantly across gender and race.  As a result, the outputs of emotion recognition models are often highly correlated with these secondary demographic signals~\cite{chaplin2015gender,soto2009emotion}.  We design approaches to first measure leakage and then to counteract this leakage.
We measure privacy in two ways: 1) using a privacy metric, which we define as the incapability of an attacker to recover demographic information from representations, and 2) by determining an adversary's ability to perform membership identification~\cite{li2013membership}, defined as the ability to determine if a given user was in a dataset from which the emotion recognition models were trained (this can be harmful if the training data are collected in a sensitive context, such as counselling or therapy).  We ask the following seven questions:
\begin{enumerate}
\itemsep-0.2em 
    \item Does demographic leakage differ in umimodal and multimodal emotion recognition models?
    \item How does the privacy metric change when a network is trained to preserve privacy?
    \item How does emotion recognition performance change when networks are trained to preserve privacy?
    \item How does the  adversarial component's strength impact emotion recognition performance and the privacy metric?
    \item Focusing on gender, how does the performance of emotion recognition change when a network is trained to preserve privacy?
    \item Does the location of the adversarial component within a network  affect the privacy metric and emotion recognition performance?
    \item Does the privacy preserving paradigm help defend against other attacks such as membership identification?
    
\end{enumerate}


Our results show that representations obtained for emotion recognition can be exploited by an adversary to predict sensitive variables given unimodal information (either audio or lexical). We further show that multimodal models contain even more sensitive information, as lexical and audio each encode different aspects of demographic information. We show that we can increase the defense against this attack by adversarially training representations to be invariant to gender. The novelty of this work is two fold: (1) we analyze how the demographic privacy of a representation differs across modalities and how it can be increased using adversarial paradigm; and, (2) we obtain privacy enhanced representations that defend against multiple privacy attacks while still maintaining performance on emotion recognition.

\section{Related Work}
\label{sec:relatedwork}
Previous research has investigated methods to improve privacy in data collection. Early work focused on rule-based systems, which would identify patterns in text and replace them with random word tokens~\cite{gomez2010data}. Other methods included the addition of background noise
or 
randomizing the order of sentences~\cite{evfimievski2002randomization}. These systems, though easy to interpret, are harder to scale to larger or varying distributions of datasets for they might necessitate an increase in the number of rules required and require expert input. 

Recent research has examined privacy preservation in the context of neural networks. These efforts have primarily focused on ensuring that the input data are not memorized and cannot be retrieved given a deployed model. 
~\cite{carlini2019secret} showed attackers could extract unique and secret sequences such as credit card numbers given models that are trained without accounting for unintended memorization. 
~\cite{abadi2016deep} proposed adding random noise to either the aggregated dataset or to individual datapoints to defend against membership query attacks. This method though, is usually either used for structured data or images and often incurs a cost in terms of a reduction in task performance.

Another line of work concentrates on fair algorithmic representation. Though the end goal isn't privacy, the methodology is similar. The aim is to create networks that are invariant to particular attributes, usually demographic information
to obtain debiased word embeddings~\cite{bolukbasi2016man}, ensure fairness pairities~\cite{corbett2018measure}, and train fair hate speech classification~\cite{davidson2019racial}.

Previous research has looked at task-specific privacy preservation for a particular attribute in a dataset.  For example, ~\cite{elazar2018adversarial} investigated text-based privacy preservation for sentiment.
\cite{zhao2019adversarial} looked at minmax modelling of the utility-privacy tradeoff by classifying gender as a primary task, while masking ethnicity and age. \cite{coavoux2018privacy} looked into modelling privacy by declustering representations that fall under the same sensitive attribute subgroup.


Given most of the previous work on privacy preserving representations concentrates on just lexical information, we tackle the questions that arise from desiring privacy preservation in multimodal representations for emotion recognition. While the primary goal of most previous works has been to avoid unintentional inference by the application itself, we concentrate on minimizing the potency of an attacker to deliberately recover sensitive attributes from an invariant representation.

\section{Datasets and Features}

\subsection{Datasets}
We use four common emotion recognition datasets:
MSP-Improv~\cite{busso2017msp}, 
MSP-Podcast~\cite{lotfian2017building},
Interactive Emotional Dyadic MOtion Capture (IEMOCAP) dataset~\cite{busso2008iemocap}, and
Multimodal Stressed Emotion (MuSE) dataset~\cite{jaiswal2019muse}. 

\textbf{MuSE.} The MuSE dataset was collected to understand the interplay between stress and emotion in natural spoken communication. It contains audio, video, thermal, physiological data and associated manual transcriptions.
The dataset consists of 55 recordings from 28 participants, for a total of 2,648 utterances. For these recordings, emotion in the participant was induced via emotionally evocative monologue topics~\cite{aron1997experimental}. Data selection was performed to reduce the dataset to include only utterances of length $[3,25]$, inline with previous emotion datasets~\cite{Khorram2018}.  

\textbf{IEMOCAP.} The IEMOCAP dataset was created to explore the relationship between emotion, gestures, and speech. Pairs of actors, one male and one female (five males and five females in total), were recorded over five sessions (either scripted or improvised)
The data were segmented by speaker turn, resulting in a total of 10,039 utterances (5,255 scripted turns and 4,784 improvised turns). It contains audio, video, and associated manual transcriptions.

\textbf{MSP-Improv.} The MSP-Improv dataset was collected to capture naturalistic emotions from improvised scenarios while partially controlling for lexical content.
The data of 8438 sentences were divided into 652 target sentences, 4,381 improvised turns (the remainder of the improvised scenario, excluding the target sentence), 2,785 natural interactions (interactions between recordings of the scenarios), and 620 read sentences (emotional readings of the target sentences).  It contains audio, video, and transcriptions derived from automatic speech recognition (ASR).

\textbf{MSP-Podcast.} The MSP-Podcast dataset was collected to build a naturlisitic emotionally balanced speech corpus by retrieving emotional speech from existing podcast recordings. This was done using machine learning algorithms, which along with a cost-effective annotation process using crowdsourcing, led to a vast and balanced dataset. We use a pre-split part of the dataset which has been identified for gender of the speakers which comprises of 13,555 utterances.  The dataset as a whole contains audio recordings.

\subsection{Labels}

\textbf{Emotion Labels.}  Utterances in each of the four datasets were labeled using the dimensional descriptions of activation (calm vs. excited) and valence (positive vs. negative).  Each utterance in the MuSE dataset was labeled on a nine-point Likert scale by eight crowd-sourced 
annotators~\cite{jaiswal2019muse}, who observed the data in random order across subjects. We average the annotations to obtain a mean score for 
each utterance, and then bin the mean score into one of three classes, defined as, \{``\textit{low}'': [min, 4.5], 
``\textit{mid}'': (4.5, 5.5], ``\textit{high}'': (5.5, max]\} valence and activation. 
Utterances in IEMOCAP and MSP-Improv were annotated on a five-point Likert scale.
The activation and valence values for were averaged over all annotations for an utterance and binned as: \{``\textit{low}'': [1, 2.75], ``\textit{mid}'': (2.75, 3.25], ``\textit{high}'': (3.25, max]\}. The labels for MSP-Podcast were annotated on a seven point Likert scale and averaged over all annotations for an utterance and binned as: \{``\textit{low}'': [1, 3.75], ``\textit{mid}'': (3.75, 4.25], ``\textit{high}'': (4.25, max]\}.

\subsection{Features}

\textbf{Acoustic.}  We use Mel Filterbank (MFB) features, which are frequently used in speech  processing applications, including speech recognition, speaker recognition, and emotion recognition.
We extract the 40-dimensional MFB features using a 25-millisecond Hamming window with a step-size of 10-milliseconds. Each utterance is represented as a 40-dimensional time series. We $z$-normalize the obtained acoustic features by speaker.
\\\textbf{Lexical.} We use the word2vec representation~\cite{mikolov2013distributed} based on the transcriptions for MuSE and IEMOCAP, which has shown success in sentiment and emotion analysis tasks.  We do not use MSP-Improv due to errors in ASR transcription or MSP-Podcast due to the lack of transcripts.
We represent each word in the input as a 300-dimensional vector 
using a pre-trained word2vec model, replacing out-of-vocab
words with the $\langle unk\rangle$ token.
Each utterance is represented as a sequence of 300-dimensional feature vectors. 

\section{Experimental Setup}
In this section, we describe the network architecture, the training recipe, and the metrics in consideration.

\subsection{Architecture}
\label{sec:setup_arch}

\begin{figure}[t]
  \centering
  \includegraphics[width=0.7\linewidth]{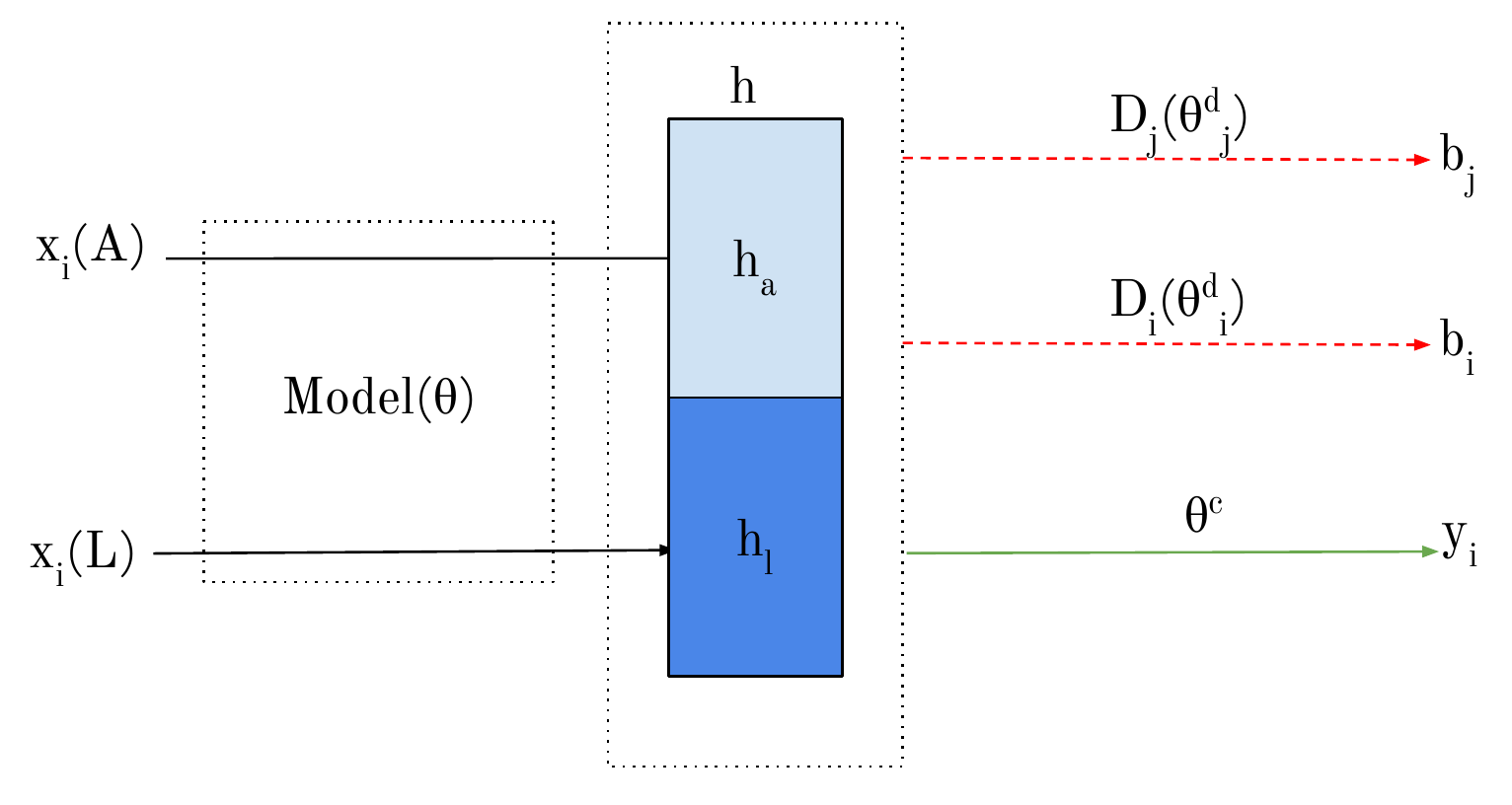}
  \caption{Privacy preserving network architecture.}
  \label{fig:model}
\end{figure}

The objective of the this system is to maximize the performance of the emotion classifier while minimizing the performance of the gender classifier (see Figure~\ref{fig:model}).
The main network consists of three components: (1) embedding sub-network, $Model(\theta)$; (2) emotion classifier, $\theta^c$ and output $y_i$; and (3) gender classifier, $D_i(\theta^d_i)$, with output, $b_i$. We then disucss how an attacker network could maliciously use this information to obtain sensitive demographic information.

\textbf{Main Network. }The \textit{embedding sub-network} uses a state-of-the-art multimodal approach in emotion recognition~\cite{aldeneh2017pooling} in which the acoustic and lexical information are processed separately and then joined after the application of modality-specific global mean pooling.  The acoustic input stream $(x_i(a))$, where $i$ represents an input frame (40-dimensional) and $a$ represents acoustic, is processed using a set of convolution layers and Gated Recurrent Units (GRU), which are fed through a mean pooling layer to produce an acoustic representation $(h_a)$. The lexical input $(x_i(l))$,  where $i$ represents an input word (300-dimensional) and $l$ represents lexical, is passed through GRUs and pooled to obtain a lexical representation $(h_l)$. For the multimodal setup, these two representations, $(h_a)$ and $(h_l)$, are concatenated $(h)$. The representations ($h, h_a, h_l$) are of fixed-length given acoustic and lexical inputs. 
The \textit{emotion classifier} takes in the representation ($h, h_a,$ or $h_l$) as input and estimates valence or activation using a set of dense layers. The \textit{gender classifier} estimates the gender label (i.e., male or female) using a set of dense layers.

\textbf{Gender-Leakage.}
The main network is trained to unlearn gender. To achieve this goal, we use a Gradient Reversal Layer (GRL)~\cite{ganin2014unsupervised}. GRLs are a common multi-task approach to train networks such that they are invariant to specific properties~\cite{meng2018speaker}.
During the backward pass of the training phase, the GRL multiplies the backpropogated gradients by $-\lambda$ (i.e., the strength of the adversarial component). During the forward pass, the GRL acts as an identity function. To make the network invairant to gender, we place the GRL function between the embedding sub-network and the gender classifier. We obtain gender-invariant representations using the following loss function:
\[\widehat{\theta}= \underset{\theta_{M}}{min}\underset{\{\theta_{D^i}\}^N_{i=1}}{max}\chi(\widehat{y}(x;\theta_M),y)-\sum_{i=1}^{N}(\lambda_i.\chi(\widehat{b}(x;\theta_{D^i}),b_{i}))\] 
\noindent where $N$ is the number of targeted sensitive variables (here $N$=$1$). The loss function ensures that while the output components are trained to be good predictors, the  representation is trained to be maximally good for the primary task (emotion) and maximally poor for the secondary task (gender).


\textbf{Attacker Network.}
We assume that the attacker has access to a held-out dataset (either a different dataset or a section of the original dataset) with known gender labels. The attacker generates representations for this dataset using the previously described embedding sub-network. 
The network then learns to predict gender label from the generated representations using a set of dense layers.  Since the parameters used to construct the representation are fixed, the attacker only acts upon its own parameters to optimize the gender prediction linear loss. 
The purpose of the attacker's network is to recover gender information from representations whose labels are unknown. Though testing using a singular network isn't a guarantee of robustness of the representation to privacy attacks, for the scope of this paper, we use a feed forward network, one of the powerful learning methods on a fixed size static representation.

\textbf{Model Variations.} We use 12 variants of the network shown in Figure~\ref{fig:model}.
We train combinations of the following setups:
\textit{\{general-classification-model (Gen), privacy-preserving-classification-model (Priv)\}} $\times$ \textit{\{activation, valence\}} $\times$
\textit{\{uni-lexical, uni-acoustic, multimodal\}}.

The general classification setup makes use of the embedding sub-network with text, speech, or both as input streams and the emotion classifier.  The privacy preserving classification setup adds the gender classifier to the general setup.

\textbf{Training}
We implement models using the Keras library.
We use a weighted cross-entropy loss function for each task and learn the model parameters using the RMSProp optimizer.
We train our networks for a maximum of 50 epochs and monitor the validation loss for the emotion classifier after each epoch, stopping the training if the validation loss does not improve for five consecutive epochs.
Once the training ends, we revert the network's weights to those that
achieved the lowest validation loss on the emotion classification task. 
For the privacy preserving classification model, we ensure that the chosen model yields a chance unweighted average recall (UAR) for the gender classification task on the validation set. 
Finally, we train each setup three times with different random seeds and average the predictions over these runs to reduce variations due to random initialization.

We use validation samples for hyper-parameter selection and early stopping. 
The hyper-parameters that we use for the main network include:
num. of convolutional layers \{3, 4\}, 
width of the convolutional layers \{2, 3\}, 
num. of convolutional kernels \{32, 64, 128\}, 
num. of GRU layers  \{2, 3\}, 
GRU layers width  \{32\}, 
num. of dense layers  \{1, 2\}, 
dense layers width  \{32, 64\}, 
GRL $\lambda$  \{0.3, 0.5, 0.75, 1\}.
For the adversarial emotion classification setups, we use the hyper-parameters that maximize the validation emotion classification performance while minimizing the validation gender classification performance.
For the attacker's model, we use the following hyper-parameters:
number of dense layers  \{2, 3, 4\}, 
dense layers width  \{32, 64\}.
We report the UAR performance of our models, given the imbalanced nature of our data.
\subsection{Metrics}
\label{subsec:metrics}
  \textbf{Performance.} We define performance for emotion recognition as the ability of the model to correctly classify either activation or valence into 3 categories: low, medium, and high.  We measure performance using UAR (chance is 0.33).
  \\\textbf{Demographic Leakage.} Leakage is defined as the ability of a trained gender classifier to predict gender from the representations which are obtained when the network is simultaneously trained to perform the primary task.
  \\\textbf{Demographic Privacy Metric.} We define privacy metric as the inability of an attacker to be able to recover gender from the representations trained on a primary task. To test this, we use four phase training.
  \begin{enumerate}
  \itemsep-0.2em 
    \item 
    We train the main network on a dataset (D1) represented by the pair $(x_{D1},y_{D1})$ where $x$ is the data input while $y$ is the gender label. We obtain representations for this dataset ($h(x_{D1})$).
    \item 
    We consider that the attacker has access to another dataset or unused subset of the same dataset (D2) represented by the pair $(x_{D2},y_{D2})$. We generate representations $h(x_{D2})$ for the pairs in this dataset using the embedding sub-network of the main network.
    \item 
    We train a model $(M_{att})$ to predict gender labels using the representations obtained in step 2, represented as $M_{att}((h(x_{D2}), y_{D2}))$.
    \item 
    Using the model obtained previously $(M_{att})$, we choose $h(x_{D1})$ as inputs, and measure the gender prediction capability of the attacker $UAR(M_{att}((h(x_{D1}), y_{D1}))$. The Demographic Privacy Metric of an attacker is then quantitatively defined as $1-UAR(M_{att})$.
  \end{enumerate}
The range of privacy metric goes from $0$ (the attacker is always correct) to $0.5$ (the attacker has a chance UAR). 
  \\\textbf{Membership Identification.} Membership identification is the possibility of an attacker being able to recognize if a speaker belongs to the training set. 
  We assume that the adversary can obtain samples from a speaker from the same distribution as that for the training set.
  Consider that the adversary knows some speakers for whom representations definitely exist in the training set and some for whom they definitely don't. We test the possibility of membership identification using four steps:
  
  \begin{enumerate}
  \itemsep-0.2em 
    \item We simulate the above using cross-validation. Given five speaker independent folds, we use three for the training set. From the remaining two folds, we add some samples of the selected speakers to the training set. 
    \item Consider the attacker knows the speakers selected and not selected for training from set four $(s4)$, but has no information about this split for set five $(s5)$. The objective of the attacker is to predict whether speakers were selected for inclusion in the training set from $(s5)$.
    \item The attacker trains a binary classification model comprised of dense layers ($M_{att-mi}$) using the representations obtained from dataset D1 as $M_{att-mi}(h(x_{D1}),`Yes')$. It obtains representations of the samples not used in training for the the selected speakers included in the training set and trains its model as $M_{att-mi}(h(x_{s4_selected}),`Yes')$ and for the speakers not included in training as $M_{att-mi}(h(x_{s4_selected}),`No')$. A speaker is saved from each label for validation.
    \item We then define the UAR of the performance of $M_{att-mi}(h(x_{s5}))$ as membership identification.
  \end{enumerate}

\section{Analysis}
In all the tables, \textbf{U} is the unweighted average recall (UAR), and \textbf{U(M)} and \textbf{U(F)} represent the performance of the model for emotion recognition when gender is male and female respectively. Leakage in the model is shown by \textbf{L}, the lower the better, where chance leakage is 0.5. Privacy metric, represented by \textbf{P} ranges from $[0,0.5]$, and is the incapability of an attacker to obtain demographic information from the representation, the higher the better.
Membership identification represented by \textbf{MI} ranges from $[0.5$ (chance UAR),$1]$, and is the capability of an attacker to identify if the subject belongs in the training set, for which the lower the value, the better. We code identify the datasets as follows: \textbf{Imp}-MSP-Improv; \textbf{Pod}-MSP-Podcast; \textbf{Iem}-IEMOCAP; and \textbf{MuS}-MuSE. All significance tests are paired t-test, with significance established (shown in bold) when \textbf{Benjamini-Hochberg adjusted}($\alpha = 0.05$) $p$-value$<0.05$.

\subsection{Question 1}
\label{subsec:q1}
\textsc{Q:} \textit{Does demographic leakage differ in umimodal and multimodal emotion recognition models?}\\
\textsc{Hypothesis}: \textit{Multimodal representations leak more gender information than unimodal representations.}\\
Previous research has shown that different modalities have varying capabilities of capturing demographic information, such as age or gender~\cite{levitan2016automatic}.The authors showed that audio, as compared to lexical, is used more successfully to predict gender. Hence, we hypothesize that a combination of these modalities leads to an increase in the leakage of the sensitive variable. 

We train the six setups separately for each dataset described in Section~\ref{sec:setup_arch} for activation and valence. We report the average across five-fold speaker-independent cross-validation in Table~\ref{normal-results-act} and Table~\ref{normal-results-val}. We find that:
\begin{itemize}
\itemsep-0.2em 
    \item A network trained to only recognize emotion is generally discriminative for gender as well. For instance we obtain a leakage of 0.69 when training a multi-modal network for activation and of 0.72 when trained for valence on MuSE.
    \item In unimodal systems, leakage is higher when systems are trained using only audio streams compared to lexical.
    \item Leakage of gender in learned representation is higher for multimodal systems than that for the unimodal systems for both, MuSE and IEMOCAP (the two datasets with both audio and lexical information).
\end{itemize}

Our results suggest that models that aren't explicitly trained for gender recognition, or, that don't use gender as an input feature, still learn representations that are discriminative to identify gender. This leakage is more prominent when the input stream is audio as compared to lexical, but the leakage compounds
in multimodal systems. 


\begin{table}[t]
\addtolength{\tabcolsep}{-2pt}
\caption{Results for activation prediction, Gen
: U-UAR, U(M/F)-UAR for male/female, L-leakage, P-privacy metric, MI-membership identification.
Bold show significant difference vs Tab.~\ref{adv-results-act} using paired t-test, adjusted p-value$<0.05$.
}
\begin{tabular}{llllllll}
\hline
\multicolumn{2}{l}{}                  & U(M)     & U(F)     & U        & L        & P & MI \\ \hline
\multirow{4}{*}{Audio}         & Imp   &   0.65   &  \textbf{0.62}    & \textbf{0.63}     &   0.69   & 0.35  &  0.71  \\
                               & Pod   &   0.69   &   0.70  &  0.70    &  0.71    &  0.32 &  0.73  \\
                               & Iem  &   0.66   &  0.69    &  0.67    &   0.73   &  0.30 &  0.72  \\
                               & MuS   &    \textbf{0.61}  &  \textbf{0.64}    &  \textbf{0.63}    &  0.72    &  0.33 &   0.75 \\ \hline
\multirow{2}{*}{Lexical}          & Iem   &  0.51    &  0.52    &  0.52    &  0.62    &  0.39 & 0.59   \\
                               & MuS   &   0.54   & 0.56     &  0.55    &    0.64  &  0.38 &  0.60  \\ \hline
\multirow{2}{*}{Multimodal}    & Iem   &   0.66   & 0.70     &  0.68    &    0.74  &  0.30 &  0.74   \\
                               & MuS   &  0.65    &   0.66   &  0.66    &   0.73   &  0.31 &  0.76  \\ \hline
\end{tabular}
\label{normal-results-act}
\end{table}

\begin{table}[t]
\addtolength{\tabcolsep}{-2pt}
\caption{Results for valence prediction, Gen
\\Bold show significant difference vs Table~\ref{adv-results-val} using paired t-test, adjusted ($\alpha = 0.05$) p-value$<0.05$.
}
\begin{tabular}{llllllll}
\hline
\multicolumn{2}{l}{}                  & U(M)     & U(F)     & U        & L        & P & MI \\ \hline
\multirow{4}{*}{Audio}         & Imp   &  0.53    &  0.49    &   \textbf{0.51}   &   0.56   &  0.44 &  0.70  \\
                               & Pod   &  0.56    &  0.57    &  0.56    &   0.60   &  0.42 &  0.71  \\
                               & Iem  &   0.60   &  0.61    &   0.60   & 0.62      &  0.39 &  0.70  \\
                               & MuS   &   0.50   &  0.47    &  0.48    &  0.58    & 0.42  &  0.72  \\ \hline
\multirow{2}{*}{Lexical}          & Iem   &  0.64    &  0.65    &  0.65    &   0.61   & 0.41  &  0.62  \\
                               & MuS   &   0.68   &  0.69    &  0.68    &    0.57  &  0.45 &  0.63  \\ \hline
\multirow{2}{*}{Multimodal}    & Iem   &  0.67    &  0.71    &  0.69    &   0.68   &  0.32 &  0.70  \\
                               & MuS   &  \textbf{0.67}    &   0.66   &  \textbf{0.67}    &   0.64   & 0.38  &   0.71 \\ \hline
\end{tabular}
\label{normal-results-val}
\end{table}

\subsection{Question 2}
\label{subsec:q2}
\textsc{Q:} \textit{How does the privacy metric change when a network is trained to preserve privacy?}\\
\textsc{Hypothesis}: \textit{Representations that are gender-invariant are less prone to leakage when attacked by an adversary, leading to better privacy preservation.}\\
Previous research has shown that obtaining a representation from a model trained invariant to be to gender, age or location leads to better protection from an attacker who tries to recover this information~\cite{coavoux2018privacy}. Previous research~\cite{elazar2018adversarial} has also shown that while the representations might be trained such that leakage of sensitive variable is reduced to chance, the attacker might still be able to recover this information. Hence, we concentrate on using this incapability as our primary metric.
To test our hypothesis, we train the adversarial variants of the six models as mentioned above, while making sure that the leakage in the models is reduced to chance performance and compare our results to those in Table~\ref{normal-results-act} and Table~\ref{normal-results-val}. We train the multimodal models only for MuSE and IEMOCAP. 
Our results in Table~\ref{adv-results-act} and Table~\ref{adv-results-val} show that:
\begin{itemize}
\itemsep-0.2em 
    \item The privacy metric is always higher when the representations are trained adversarially, compared to generally.
    \item Even when leakage is adversarially reduced to chance, the attacker is still able to recover information about gender.
    \item The privacy metric is in general always lower for audio than for lexical based unimodal systems.
    \item Multimodal systems often have the lowest privacy metric.
\end{itemize}
Our results suggest that, though the privacy of the learned representation is improved by reducing leakage while training, the attacker can still recover that information. This effect is especially compounded for multimodal systems. While previous work has concentrated on text (Section ~\ref{sec:relatedwork}), our work shows how audio is the major culprit and that models involving audio as input are easier to exploit, even when trained adversarially for privacy preservation.

\begin{table}[t]
\addtolength{\tabcolsep}{-2pt}
\caption{Results for activation prediction, Priv
\\Bold show significant difference vs Table~\ref{normal-results-act} using paired t-test, adjusted ($\alpha = 0.05$) p-value$<0.05$.
}
\begin{tabular}{lllllll}
\hline
\multicolumn{2}{l}{}                  & U(M)     & U(F)     & U        & P        & MI \\ \hline
\multirow{4}{*}{Audio}         & Imp   &  0.64    &  0.57    &   0.60   &   \textbf{0.44}   &  0.68     \\
                               & Pod   &  0.68   &  0.69    &   0.69   &   \textbf{0.44}   &  \textbf{0.68}     \\
                               & Iem  &   \textbf{0.68}   &    0.70  &   \textbf{0.69}   &     \textbf{0.43}  &    \textbf{0.67}   \\
                               & MuS   &   0.58   &  0.61    &  0.60     &   \textbf{0.45}  &   \textbf{0.69}    \\ \hline
\multirow{2}{*}{Lexical}          & Iem   &  \textbf{0.55}    & \textbf{0.56}     &  \textbf{0.56}    &  \textbf{0.48}   &  \textbf{0.55}     \\
                               & MuS   &   \textbf{0.58}   &   0.57   & \textbf{0.58}    &   \textbf{0.47}  &   0.58    \\ \hline
\multirow{2}{*}{Multimodal}    & Iem   & 0.66     &   0.69   &  0.68    &   \textbf{0.41}   &    \textbf{0.67}   \\
                               & MuS   &   0.65   & 0.64  &  0.65    &   \textbf{0.43}   &   \textbf{0.69}    \\ \hline
\end{tabular}
\label{adv-results-act}
\end{table}

\begin{table}[t]
\addtolength{\tabcolsep}{-2pt}
\caption{Results for valence prediction, Priv
\\Bold show significant difference vs Table~\ref{normal-results-val} using paired t-test, adjusted ($\alpha = 0.05$) p-value$<0.05$.
}
\begin{tabular}{lllllll}
\hline
\multicolumn{2}{l}{}                  & U(M)     & U(F)     & U        & P        & MI \\ \hline
\multirow{4}{*}{Audio}         & Imp   &   0.51   &  0.49    &  0.48    &   \textbf{0.48}   &  0.68     \\
                               & Pod   &   0.55   &  0.56     &   0.56   &   \textbf{0.47}   &    0.70   \\
                               & Iem  &   0.60   &  0.62    &   0.61   &     \textbf{0.45}  & 0.68       \\
                               & MuS   &   0.48   & 0.47     &   \textbf{0.46}   &    0.46  &  0.71     \\ \hline
\multirow{2}{*}{Lexical}          & Iem   &  \textbf{0.67}    & \textbf {0.68}     &   \textbf{0.67}   &   \textbf{0.46}   &   0.62    \\
                               & MuS   &  0.70    &   0.71   &   0.70   &   0.47   &  0.62    \\ \hline
\multirow{2}{*}{Multimodal}    & Iem   &  0.68    &   0.70   &  0.69    &   \textbf{0.45}   &  \textbf{0.68}     \\
                               & MuS   &  0.64    &  0.65    &  0.65    &   \textbf{0.46}   &    0.71   \\ \hline
\end{tabular}
\label{adv-results-val}
\end{table}

\subsection{Question 3}
\label{subsec:q3}
\textsc{Q:} \textit{How does emotion recognition performance change when networks are trained to preserve privacy?} \\
\textsc{Hypothesis}: \textit{There is a minor drop in emotion recognition performance when models are trained to preserve privacy.}\\
Previous research has shown that training a model invariant to a dataset variable might lead to drop in performance on the primary task, especially when there exists known correlations or biases in the datasets between the target label for the primary task and the secondary task~\cite{meng2018speaker} 

We compare the performance for predicting activation and valence of the models trained just to predict emotion (Act: Table~\ref{normal-results-act}, Val: Table~\ref{normal-results-val}) versus the model trained to enhance privacy while still predicting emotion in (Act: Table~\ref{adv-results-act}, Val: Table~\ref{adv-results-val}).
Our results suggest that, in general there is no significant effect on the performance on primary task when we train privacy preserving networks. We find that the performance is either maintained, e.g., Act: multimodal-MuSE; Val: multimodal-IEMOCAP, or there is a slight decrease in performance for some setups, e.g., Act: unimodal-acoustic-MuSE; Val: multimodal-MuSE. In multiple cases, such as Act/Val:unimodal-lexical-MuSE/IEMOCAP, contrary to some previous work, we also see a significant increase in performance, implying that making the model invariant to gender increases its robustness by not learning replicable associations between gender and emotion label. 

\subsection{Question 4}
\uline{Q:} \textit{How does the  adversarial component's strength impact emotion recognition performance and the privacy metric?}\\
\uline{Hypothesis}:\textit{As the strength of the adversarial component increase, the privacy metric increases and the performance on the pimary task is unchanged.}\\
Our results in Section~\ref{subsec:q2} suggest that while the leakage of the model was reduced to chance performance, the attacker is still capable of recovering this information. We analyze the effect of the strength of the adversarial component on the performance of the primary task and the privacy metric. 

We find that the emotion recognition performance is generally unaffected with change in $\lambda$, as expected from results in Section~\ref{subsec:q3}. 
We observe that the the attacker is usually less capable of inferring gender from learned representations when $\lambda=0.50$ as compared to when $\lambda=0.75$. For example, the privacy metric for unimodal audio system trained on MuSE increases from $0.39$ to $0.45$. But contrary to our expectation, we often see a decrease in the privacy metric when we move from $\lambda=0.75$ to $\lambda=1.00$ for both activation and valence. For example, the privacy metric for the unimodal audio system trained on MuSE decreases from $0.45$ to $0.41$. 
 The decrease in the privacy metric as $\lambda\rightarrow1$ could be attributed to overfitting of data~\cite{schmidt2018adversarially} when being trained for invariance to the sensitive variable which the attacker network is able exploit.
This suggests that an increase in the strength of the adversarial component doesn't necessarily correlate to an increase in the privacy metric.

\subsection{Question 5}
\textsc{Q:} \textit{Focusing on gender, how does the performance of emotion recognition change when a network is trained to preserve privacy?}\\
\textsc{Hypothesis:} \textit{Learning representations invariant to gender will affect performance on the primary task in an imbalanced manner across subgroups.}\\
Previous research~\cite{bagdasaryan2019differential} has shown that training models invariant to race or gender can harm performance for one group more than others. This may be worrying when the prediction is used for sensitive application such as intervention or policing. Hence, we analyze if the performance on emotion recognition is affected in an imbalanced way for the models trained to enhance privacy. 

We compare the performance for predicting activation and valence of the models trained just to predict emotion (Act: Table~\ref{normal-results-act}, Val: Table~\ref{normal-results-val}) versus the model trained to enhance privacy while still predicting emotion in (Act: Table~\ref{adv-results-act}, Val: Table~\ref{adv-results-val}). We find that while the performance is affected differently for the subgroups, the effect is not consistent across multiple setups and datasets. For example, the unimodal-acoustic system trained on MSP-Improv for activation classification decreases in performance for both male and female groups, but the effect on female group is greater. But the pattern isn't consistent across other datasets for the same model setup. Our takeaway from this analysis is cautionary, that though the privacy metric increases when a model is adversarially trained to enhance privacy, we need to ensure that the performance of the model on that dataset doesn't harm one subgroup more than the other.

\subsection{Question 6}
\textsc{Q:} \textit{Does the location of the adversarial component within a network  affect the privacy metric and emotion recognition performance?}\\
\textsc{Hypothesis:} \textit{Unlearning the demographic variable in separate pooled streams will improve the privacy metric.}\\
Previous work has shown that curtailing a variable on intermediate layers often leads to a difference in the performance of the classifier~\cite{chabanne2017privacy}. As seen in section~\ref{subsec:q1}, audio is more prone to leakage than lexical, hence, a multimodal system's privacy metric might benefit from curtailing audio separately. Our initial multimodal model (Fig~\ref{fig:model}) only allows for the same strength and parameters of the adversarial component to be applied for both audio and lexical streams. 
To test our hypothesis, we place the same adversarial component after the mean pooling layer of both input streams, allowing us separate control of gender invariance for both modalities, before concatenation of representation.
Our results are as follows (Bold show significant difference vs Table~\ref{adv-results-act} and Table~\ref{adv-results-val} using paired t-test, adjusted ($\alpha = 0.05$) p-value$<0.05$.):
\begin{enumerate}
\itemsep-0.2em 
    \item Act: [Iem - U: 0.66, P: \textbf{0.43}], [MuS - U: 0.63, P: 0.44]
    \item Valence: [Iem - U: 0.67, P: \textbf{0.46}], [MuS - U: 0.64, P: 0.46]
\end{enumerate}
We find that, using adversarial component separately for each input stream improves privacy metric for emotion recognition models trained on both datasets, as compared to using one adversarial component. This suggests that a granular control of invariance over modalities leads to better defense of representations against gender identification.


\subsection{Question 7}
\textsc{Q:} \textit{Does the privacy preserving paradigm help defend against other attacks such as membership identification?}\\
\textsc{Hypothesis:} \textit{Membership identification will decrease when models are trained to be invariant to speaker.}\\
Membership identification is defined as an attack that tries to identify if samples from a speaker `x' were present in the training set~\cite{li2013membership}. ~\cite{papernot2018marauder} showed that removing identifying factors from learned representations reduces the probability of membership leakage. 
For this analysis, we ask two questions: 
(a) can we defend against membership identification using a proxy task?; and
(b) can we defend against both, gender and membership identification?

We train an attack model for membership identification as specified in Section~\ref{subsec:metrics}. We find that while adversarial removal of gender in the learned representation (Act: Table~\ref{adv-results-act} and Val: Table~\ref{adv-results-val}) does lead to reduced membership identification, as compared to a model trained solely for emotion recognition (Act: Table~\ref{normal-results-act} and Val: Table~\ref{normal-results-val}), the membership identification is still far higher than chance.

Our goal is to be unable to identify whether samples from speaker `x' exist in the training set.
This is different from the usual membership defense that prevents prediction of presence of a data-point pair $(input_x,output_x)$ is in the training set. As a result, we require a proxy task, 
because our model cannot use samples from the speakers not in the training set even to induce invariance. We hypothesize that given randomly chosen speakers from the population, speaker-invariant training leads to representations that are less likely to encode speaker-specific information. This will make it harder for the attacker to identify membership of a particular speaker in the training set. 
We train the emotion recognition models specified in Section~\ref{sec:setup_arch} and replace the gender invariance sub-network with speaker invariance and use the same membership attack network. Our results are as follows (Bold show significant difference vs Table~\ref{adv-results-act} and Table~\ref{adv-results-val} using paired t-test, adjusted ($\alpha = 0.05$) p-value$<0.05$.):
\begin{enumerate}
\itemsep-0.2em 
    \item Activation: Unimodal-Acoustic (Imp - \textbf{0.58}, Pod - \textbf{0.60}, Iem - \textbf{0.58}, MuS - \textbf{0.62}); Unimodal-Lexical (Imp - 0.52, MuS - 0.53); Multimodal (Imp - \textbf{0.58}, MuS - \textbf{0.60})
    \item Valence: Unimodal-Acoustic (Imp - \textbf{0.54}, Pod - \textbf{0.56}, Iem - \textbf{0.59}, MuS - \textbf{0.60}); Unimodal-Lexical (Imp - \textbf{0.52}, MuS - \textbf{0.53}); Multimodal (Imp - \textbf{0.57}, MuS - \textbf{0.58})
\end{enumerate}

We find that models trained to be invariant to speaker identity have significantly lower UAR for membership identification than those trained solely to recognize emotion, or trained invariant to gender, which matches our hypothesis.

\textbf{Extension towards multi-attribute invariance}
We train our emotion recognition model using both the adversarial components (speaker id and gender) and the primary classification task i.e. emotion recognition. This ensures that the model can defend against both, gender and membership identification attacks. We report our results in Table~\ref{adv-multi-act-val}. We find that we can successfully train models that are safer against both, gender and membership identification attacks, while still maintaining similar performance on the primary task, as an evidence towards multi-attribute invariance.

\begin{table}[]
\addtolength{\tabcolsep}{-2pt}
\caption{Results for activation and valence prediction, when adversarially unlearning both subject identity and gender
\\Bold show significant difference vs Table~\ref{normal-results-act} and Table~\ref{normal-results-val} using paired t-test, adjusted ($\alpha = 0.05$) p-value$<0.05$. Italic signifies that the performance has significantly decreased.
}
\begin{tabular}{llllllll}
\hline
                            &    & \multicolumn{3}{c}{Activation} & \multicolumn{3}{c}{Valence} \\ \cline{3-8} 
                            &    & U          & P         & MI     & U       & P       & MI       \\ \hline
\multirow{4}{*}{Audio}         & Imp   &   \textit{0.59}   &  \textbf{0.45}   &  \textbf{0.58} 
& \textit{0.47}    & 0.48  &  \textbf{0.53}  \\
                               & Pod   &  0.69    & \textbf{0.46}     &  \textbf{0.59}    &   0.54   & \textbf{0.48}  &  \textbf{0.56} \\
                               & Iem  &  0.67    &  \textbf{0.43}    &  \textbf{0.57}    &  0.60    &  \textbf{0.47} &   \textbf{0.57} \\
                               & MuS   & \textit{0.59}      &   \textbf{0.44}   &   \textbf{0.60}   &  \textit{0.46}    & \textbf{0.46}   &  \textbf{0.58}  \\ \hline
\multirow{2}{*}{Lexical}          & Iem   &    0.53  &   \textbf{0.48}   &   \textbf{0.52}   &  0.66    & \textbf{0.47}  &  \textbf{0.53}  \\
                               & MuS   &  0.54      &    \textbf{0.47}  &  \textbf{0.52}    &   0.68   & 0.46   &  \textbf{0.53}  \\ \hline
\multirow{2}{*}{Multimodal}    & Iem   &  0.66    & \textbf{0.40} &   \textbf{0.57}   &   0.65   & \textbf{0.43}  &  \textbf{0.56}  \\
                               & MuS   &   0.65   &   \textbf{0.40}   &   \textbf{0.58}   &  \textit{0.62}    & \textbf{0.44}  &   \textbf{0.58} \\ \hline
\end{tabular}
\label{adv-multi-act-val}
\end{table}


\section{Conclusion}
In this work, we show how privacy preserving networks trained for emotion recognition can be used to protect against gender and membership identification. 
This provides a compelling case for separating the process of data processing on user devices and of task-specific training on central servers, thus increasing the privacy of the user. 
While in this paper we concentrate on a single primary task i.e. emotion recognition for this paper, this method can be extended to maximize utility on multiple primary tasks that are loosely related to each other and are benefited from a multi-task setup as shown for dialogue act and turn detection, and sentiment and topic classification~\cite{ruder2017overview}. 

For future work, we aim to explore how privacy enhanced representations can be learned for multiple primary tasks such as speaker verification and emotion recognition that may not be related to each other. This would enable us to deploy a generalized privacy model in form of SaaS which all developers could use the to obtain privacy enhanced representations that are then stored on the central server.

\bibliography{aaai.bib}
\bibliographystyle{aaai}

\end{document}